\documentclass[sigconf]{acmart}

\usepackage{booktabs}
\usepackage{multirow}
\usepackage{hyperref}
\usepackage{url}
\usepackage{graphicx}
\usepackage{enumitem}
\usepackage{amsmath}
\usepackage{algorithm}
\usepackage{algpseudocode}
\usepackage{colortbl} 
\usepackage{xcolor}   
\usepackage{multirow}
\usepackage{rotating} 
\usepackage{array} 
\usepackage{graphicx} 
\usepackage{lipsum} 
\usepackage{caption} 
\usepackage{wrapfig}
\usepackage{soul}
\usepackage{float}
\usepackage{bm}
\usepackage{placeins}

\AtBeginDocument{%
  }

\setcopyright{none}
\renewcommand\footnotetextcopyrightpermission[1]{}
\begin{document}

\title{Multi-Ontology Integration with Dual-Axis Propagation \\
 for Medical Concept Representation}

\author{Mohsen Nayebi Kerdabadi}
\affiliation{%
  \institution{University of Kansas}
  \city{Lawrence}
  \state{KS}
  \country{USA}}
\email{mohsen.nayebi@ku.edu}

\author{Arya Hadizadeh Moghaddam}
\affiliation{%
  \institution{University of Kansas}
  \city{Lawrence}
  \state{KS}
  \country{USA}}
\email{a.hadizadehm@ku.edu}

\author{Dongjie Wang}
\affiliation{%
  \institution{University of Kansas}
  \city{Lawrence}
  \state{KS}
  \country{USA}}
\email{wangdongjie@ku.edu}

\author{Zijun Yao}
\authornote{Corresponding author.}
\affiliation{%
  \institution{University of Kansas}
  \city{Lawrence}
  \state{KS}
  \country{USA}
}
\email{zyao@ku.edu}

\renewcommand{\shortauthors}{Nayebi Kerdabadi et al.}

\captionsetup[figure]{font={stretch=0.5}} 

\newcommand{\fix}{\marginpar{FIX}}
\newcommand{\new}{\marginpar{NEW}}

\newcommand{\model}{\textsc{LINKO}}
\newcommand{\vertical}{VMP}
\newcommand{\horizontal}{HMP}

\begin{abstract}

Medical ontology graphs map external knowledge to medical codes in electronic health records (EHRs) via structured relationships. By leveraging domain-approved connections (e.g., parent-child), predictive models can generate richer medical concept representations by incorporating contextual information from related concepts. However, existing literature primarily focuses on incorporating domain knowledge from a single ontology system, or from multiple ontology systems (e.g., diseases, drugs, and procedures) in isolation, without integrating them into a \textit{unified} learning structure. Consequently, concept representation learning often remains limited to intra-ontology relationships, overlooking cross-ontology connections that could enhance the richness of healthcare representations.
In this paper, we propose \model{}, a large language model (LLM)-augmented integrative ontology learning framework that leverages multiple ontology graphs simultaneously by enabling dual-axis knowledge propagation both within and across heterogeneous ontology systems to enhance medical concept representation learning.
Specifically, \model{} first employs LLMs to provide a graph-retrieval-augmented initialization for ontology concept embedding, through an engineered prompt that includes concept descriptions, and is further augmented with ontology graph relations and task-specific details. Second, our method jointly learns the medical concepts in diverse ontology graphs by performing knowledge propagation in two axes: (1) intra-ontology vertical propagation across hierarchical ontology levels and (2) inter-ontology horizontal propagation within every level in parallel.
Last, through extensive experiments on two public datasets, we validate the superior performance of \model{} over state-of-the-art baselines. As a plug-in encoder compatible with existing EHR predictive models, \model{} further demonstrates enhanced robustness in scenarios involving limited data availability and rare disease prediction.\footnote {The code is available at \footnotesize\url{https://github.com/mohsen-nyb/LINKO.git}.}

\end{abstract}



\maketitle

\section{Introduction}
\label{introduction}

With the ubiquity of electronic health records (EHRs) in modern healthcare systems, developing machine learning models to analyze comprehensive medical histories has shown great potential in enhancing a wide range of predictive tasks \citep{choi2016doctor, poulain2024graph, jiang2023graphcare, moghaddam2024contrastive, nayebi2023contrastive, hadizadeh2025discovering}. Yet, the inherent complexity of medical concepts, i.e. EHR codes, characterized by diversity, sparsity, and temporal variability, poses a challenge for learning expressive and robust representations, particularly for infrequent codes, such as rare diseases.

A promising direction to this challenge is to incorporate domain knowledge into the concept representation learning, particularly the structured medical knowledge provided by ontology graphs. By capturing rich domain context of medical codes and their interrelationships, ontologies serve as valuable knowledge bases that support deeper interpretation of codes in EHRs~\cite{who_icd}. Typically, these ontologies offer hierarchical classifications that represent medical concepts from general to specific definitions. Leveraging these hierarchies enables learning algorithms to better capture associations among medical codes, yielding more robust representations and more accurate predictions of rare concepts.

Accordingly, recent research has placed growing emphasis on augmenting EHR representations through the incorporation of supplementary ontology graphs~\citep{choi2017gram, shang2019pre, ye2021medpath, zhang2020hierarchical, an2023kampnet, cheong2023adaptive}. However, a critical limitation of current literature is the lack of a \emph{unified} learning framework that can effectively accommodate and integrate multiple ontology systems. Specifically, existing methods typically treat each ontology as an isolated structure, supporting only vertical message passing (e.g., parent-child concept aggregation). As a result, concept representation learning is confined to intra-ontology relationships among homogeneous concepts (e.g., parent disease to child disease). To address this, there is a need to fuse multiple ontologies into a unified heterogeneous multi-knowledge graph that supports the incorporation of cross-ontology relationships (e.g., disease-drug, disease-procedure) into healthcare representation learning. These relationships should span all levels of the hierarchy, enabling horizontal message passing both \emph{within} ontologies (e.g., parent disease to parent disease at the same level) and \emph{across} ontologies (e.g., parent disease to parent drug, or child disease to child drug). Such a unified framework can enable comprehensive utilization of rich and diverse medical knowledge bases for more effective medical concept representation learning.

Moreover, large language models (LLMs), pre-trained on extensive corpora such as biomedical literature, textbooks, and websites, provide a rich source of domain-specific knowledge that, when carefully prompted, can significantly enhance various predictive tasks. A substantial body of research highlights their potential as knowledge bases \cite{petroni2019language, jiang2023graphcare, xu2024ram, alkhamissi2022review}. However, augmenting healthcare tasks with LLMs presents a significant challenge of factual inconsistency, that LLMs may generate inconsistent or misleading content in free-text responses \cite{10.1145/3703155, zhang2023siren}. In a high-stakes domain like healthcare, relying on such outputs for decision-making poses significant risks.
This underscores the need for better strategies to leverage LLMs as expert knowledge sources in healthcare predictive modeling while ensuring reliability and minimizing risk.

To address the aforementioned gaps, we propose an integrative graph learning architecture, named \textbf{\model{}}: \textbf{L}LM-augmented \textbf{IN}tegrative \textbf{K}nowledge Propagation over \textbf{O}ntology Graphs. \model{} enables knowledge propagation across multiple ontologies along both vertical and horizontal axes. Specifically, it facilitates (1) intra-ontology concept associations through a vertical knowledge propagation within each ontology graph, and (2) inter-ontology concept interactions via horizontal knowledge propagation across diverse ontology graphs. By integrating the dual-axis message passing in a \emph{unified} structure, \model{} is advantageous in capturing both homogeneous and heterogeneous concept interactions at all levels of EHR concept granularity, from the coarsest (highest) to the finest (lowest) levels, therefore enabling a synergistic learning approach that integrates EHR mining (observational patterns) with ontology structure learning (external domain knowledge). 

Furthermore, \model{} employs LLMs through a graph-retrieval-augmented initialization, where each concept and its local ontology context (e.g., connected concepts and their descriptions) form a prompt to retrieve dense embeddings from the LLM. These embeddings, which capture prior knowledge encoded in the LLM, initialize node representations in ontology graphs and are subsequently refined through dual-axis propagation. This use of LLM-derived embeddings serves solely as an augmentation step rather than a direct decision-making component (e.g., predictions, clustering) in EHR modeling. By encoding semantic concept information and then being refined within the graph, these embeddings enhance predictive performance while reducing the risk of hallucinations typically associated with free-text LLM outputs.

Lastly, our concept encoder can function as a plug-in module to existing healthcare predictive models, enhancing concept learning and improving performance.
To validate our work, we conducted extensive experiments on two widely used EHR datasets, MIMIC-III and MIMIC-IV, performing sequential diagnosis prediction, plug-in enhancement evaluation, baseline comparisons, data insufficiency tests, and interpretative case studies. The results demonstrate that \model{} significantly improves the encoding phase of healthcare predictive models, leading to enhanced predictive performance.

\section{Methodology}
\label{method}

\begin{figure*}[t]
\setlength{\abovecaptionskip}{5pt} 
    \begin{center}
    \includegraphics[width=0.85\linewidth]{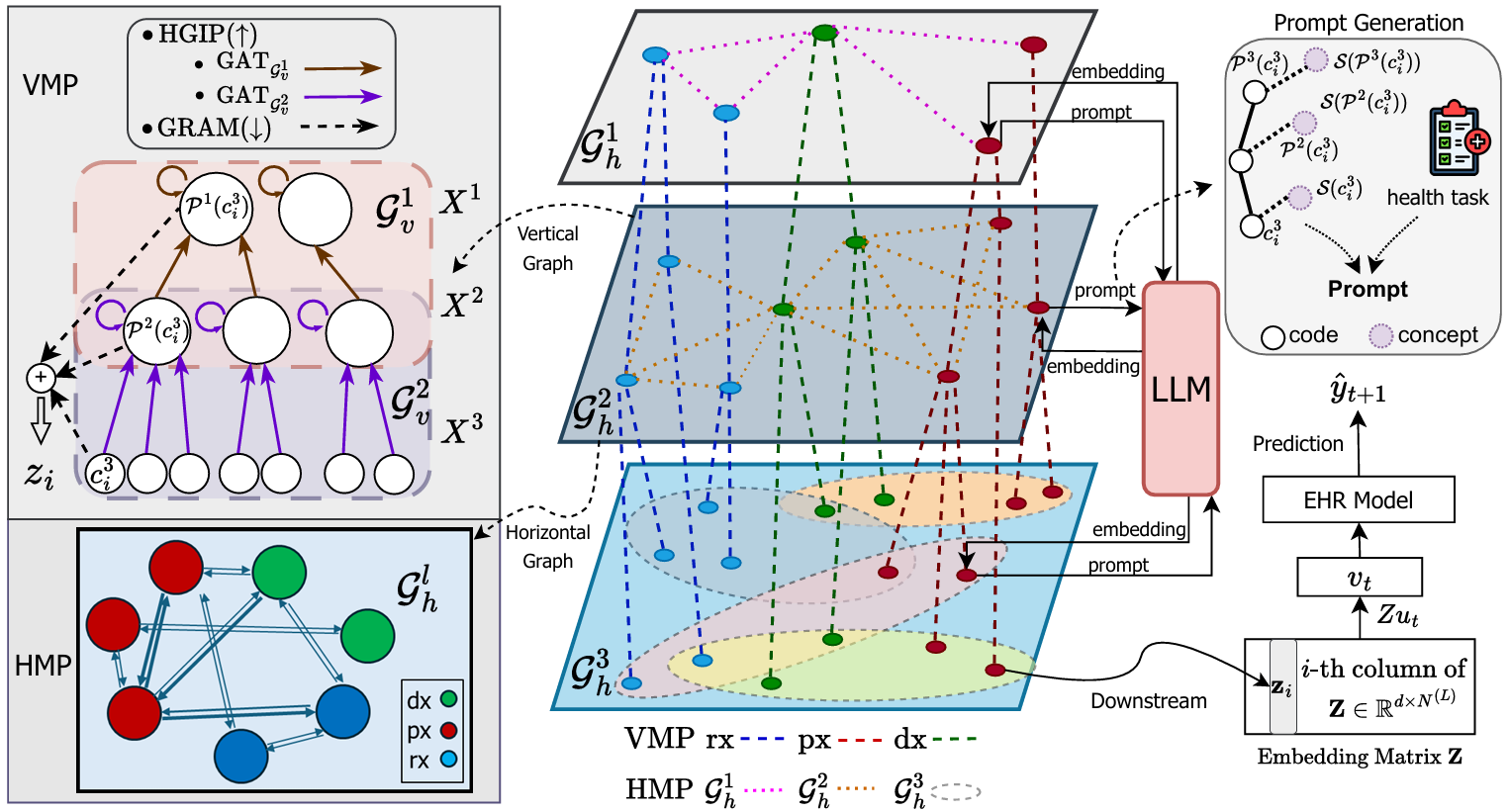}
    \caption{\model{} has three steps: (1) Meta-KG construction, which integrates heterogeneous medical concepts from multiple ontologies and initializes their embeddings using graph-augmented LLM dense vectors. The Meta-KG supports Horizontal (co-occurrence) and Vertical (hierarchical) Message Passing. (2) Horizontal Message Passing (HMP), which links concepts across ontologies at corresponding hierarchy levels based on EHR co-occurrence, implemented with either a regular graph network (e.g., GAT) or a hypergraph network (e.g., HAT). (3) Vertical Message Passing (VMP), which first performs bottom-up propagation via HGIP and then top-down propagation via GRAM. The resulting embeddings of the lowest-level medical codes are used in downstream predictive tasks.}
    \label{fig:Model}
    \end{center}
    \vspace{-0.3cm}
\end{figure*}

\subsection{Notation and Problem Definition}

\textbf{EHR Dataset.} Denoted by \(\mathcal{D} = \{\mathcal{X}_{j} \mid j \in \mathcal{J}\}\), an EHR dataset consists of the medical histories from a collection of patients \(\mathcal{J}\). For each patient, their history \(\mathcal{X}_{j}\) consists of a sequence of \(T_j\) clinical visits so that \(\mathcal{X}_{j} = \{V_{j,t}\}_{t=1}^{T_j}\). Each visit \(V_{j,t}\) is a set of \(N_t\) medical codes thereby \(V_{j,t} = \{c_{i}\}_{i=1}^{N_{t}}\), where each code $c_i$ represents a diagnosis (\textit{dx}), a prescription (\textit{rx}), or a procedure (\textit{px}). Each medical code \(c_i\) can also be associated with a descriptive name \(\mathcal{S}(c_i)\), which is typically a short text provided by the ontology. For brevity, we omit the subscripts $j$ and $t$ denoting each patient and visit. 

\vspace{0.1cm}
\noindent\textbf{Medical Ontology.} A medical ontology is a hierarchical tree-like structure that organizes clinically related concepts from general categories at the upper levels, to specific concepts at the lower levels. Each group contains clinically related codes organized under a shared parent code, representing a more general concept. For different types of medical concepts (e.g., diseases, drugs, procedures, etc.), there exists a unique ontology (e.g., ICD \cite{who_icd}, ATC \cite{who_atc}). A medical concept in an ontology is denoted as \(c^{(l)}_i \in \bm{c}^{(l)}\), where \(l \in [1:L]\) indexes the ontology level (from the highest to lowest), \(i \in [1:N^{(l)}]\) indexes the concept at the \(l\)-th level with \(N^{(l)}\) total number of code, and \(\bm{c}^{(l)}\) represents the set of all concepts at level \(l\). 
Last, we define the function \(\mathcal{P}^{k}(c_{i}^{(l)})\), which maps a concept \(c_{i}^{(l)}\) at level $l$ to its ancestor or descendant concepts at level \(k\). If \(k > l\), it returns the set of descendants; if \(k < l\), it returns the ancestor; and if \(k = l\), it returns the concept itself. 
Medical concepts (codes) in EHRs are typically located at the leaf level of the ontology (\(\bm{c} = \bm{c}^{(L)}\)).

\vspace{0.1cm}
\noindent\textbf{Healthcare Predictive Tasks.}  
Given a patient’s visit sequence \(\mathcal{X}_{j} = \{V_{1}, V_{2}, \ldots, V_{T_j}\}\), the objective is to predict clinical outcomes. These may involve binary tasks (e.g., mortality, readmission) or multi-label classification (e.g., prescription recommendation, diagnosis prediction). In this paper, we focus on the latter--predicting diagnosis codes for the next visit \(V_{T_j+1}\), a comprehensive task aimed at identifying potential diseases or conditions for future encounters.

\vspace{-3mm}

\subsection{Method Overview}
We present an overview of our proposed method, \model, a hierarchically integrative multi-knowledge graph encoder designed for \emph{medical concept} representation learning. This encoder can be added to different EHR predictive models to enhance medical concept representation and improve performance. The framework depicted in Figure \ref{fig:Model} is summarized in three key steps:

\textbf{Step 1:} Formulating the Meta-KG denoted by $\mathcal{G}$, a multifaceted knowledge graph (KG) that joins multiple ontology graphs through every level of hierarchy. The initialization of node embeddings in Meta-KG is generated through a curated LLM prompting strategy and dense embedding retrieval. This Meta-KG consists of \(L\) horizontal inter-ontology graphs, one at each level, and two vertical intra-ontology graphs per ontology (bottom-up and top-down).

\textbf{Step 2:} 
Carrying out multi-level Horizontal Message Passing (HMP) over inter-ontology concepts enabled by co-occurrence-based edges. We construct horizontal graphs in Meta-KG using two approaches: (1) regular graph structure, where edges are defined by concept co-occurrence probabilities; and (2) hypergraph structure, where each edge connects all concepts within a visit.

\textbf{Step 3:} 
Performing Vertical Message Passing (VMP) over intra-ontology concept edges derived from parent-child relationships in the ontology hierarchy. Within each ontology, the process begins with bottom-up propagation, where embeddings flow from child concepts to their parents, followed by top-down propagation, where updated embeddings are passed back to the children, yielding the final representations for leaf nodes.

\vspace{-3mm}
\subsection{Step 1: Graph-Augmented LLM Initialization} 

We construct Meta-KG nodes by collecting medical concepts across all hierarchical levels of each ontology. For each ontology, we first identify leaf nodes at the lowest level and then progressively map them to higher levels, thus defining nodes across the hierarchy. We use embedding \(x^{(l)}_{i} \in \mathbb{R}^{d}\) to represent each medical concept \(c^{(l)}_{i}\) in the Meta-KG. 
We extract the concept name associated with code \(c^{(l)}_{i}\), denoted by \(\mathcal{S}(c^{(l)}_{i})\). We develop a Graph-Augmented prompting strategy specifically tailored for EHRs to retrieve embeddings from LLMs for each concept. These embeddings are then used to initialize \( x^{(l)}_{i} \), which is further refined through end-to-end graph learning. This approach leverages the invaluable knowledge of LLMs while minimizing the risk of hallucination, as we rely solely on their embeddings rather than their final natural language responses.
We empirically found the following prompting strategy (Pr) most effective:

\begin{itemize}[leftmargin=*]
    \item \textbf{For \(l = 1\), prompt:} \textit{<< For the task of [task name], provide a semantic representation for [type] code [\(c^{(l)}_{i}\)] which represents [\(\mathcal{S}(c^{(l)}_{i})\)], a general medical concept. >> } 
    \item \textbf{For \(l > 1\), prompt:} \textit{<< For the task of [task name], provide a semantic representation for [type] code [\(c^{(l)}_{i}\)] which represents [\(\mathcal{S}(c^{(l)}_{i})\)]. It falls under the broader [type] categories of [\(\mathcal{P}^{l-1}(c^{(l)}_{i})\)] \\ ([\(\mathcal{S}(\mathcal{P}^{l-1}(c^{(l)}_{i}))\)]), …, [\(\mathcal{P}^{1}(c^{(l)}_{i})\)] ([\(\mathcal{S}(\mathcal{P}^{1}(c^{(l)}_{i}))\)]). >>}  
\end{itemize}
where ``task name'' can be any predictive task, e.g. diagnosis prediction, ``type'' refers to name of the ontology concept system (e.g., ICD-9 diagnosis/procedure, and ATC drug). For each code, we provide the code and its descriptive concept, followed by its broader categories or EHR ancestors and their descriptions, identified at higher levels of the ontology graph using the mapping function \(\mathcal{P}\). An example of an LLM prompt for a diagnosis code is as follows: \textbf{ICD-9 Diagnosis 250.7:} \textit{Prompt: << For the task of diagnosis prediction, provide a semantic representation for ICD-9 diagnosis code 250.7, which represents Diabetes with peripheral circulatory disorders. It is a specific medical concept under the broader ICD-9 Diagnosis categories of 250 (Diabetes mellitus), 249-259 (Diseases of Other Endocrine Glands), and 240-279 (Endocrine, Nutritional, and Metabolic Diseases, and Immunity Disorders).>>}

We employ the OpenAI off-the-shelf model, GPT text-embedding-3-small \citep{openai2023}, denoted as \(\mathcal{LLM}\) to generate a semantic embedding, containing clinical knowledge and context background from LLMs. We initialize the vector representation \(x^{(l)}_{i} \in \mathbb{R}^{d}\) as follows: 
\begin{equation}
    x_{i}^{(l)} = \mathcal{LLM}(\text{Prompt}(c_i^{(l)})) , \quad l = 1, \ldots, L, \quad i = 1, \ldots, N^{(l)}
\label{eq:llm}
\end{equation}

\subsection{Step 2: Horizontal Message Passing (HMP)}

We formulate horizontal information propagation that links medical concepts within and across ontologies at all levels. Leveraging observational co-occurrence patterns in EHR visits, we establish edges between frequently co-occurring concepts and apply GNN-style message passing. Horizontal graph edges at higher hierarchy levels are established by linking ancestor concepts mapped from observed codes at the lowest level.
This operation achieves three key goals: 1) hierarchically fusing diverse ontologies (diagnosis, drug, and procedure) to capture heterogeneous code interactions, 2) utilizing information at all levels of granularity for EHR representation. While leaf-level EHR codes offer detailed but sparse insights, mapping to higher-level concepts reduces the graph complexity, allowing us to leverage both fine-grained and coarse-grained information simultaneously for representation learning, and 3) enabling a unified approach that combines EHR mining (clinical statistics) with ontology-driven learning (defined expert knowledge).

To construct graph edges based on EHR co-occurrence information, we first create a leaf (child) level co-occurrence count matrix $Q^{(L)} \in \mathbb{R}^{N^{(L)} \times N^{(L)}}$, where $Q_{ij}$ denotes the number of occurrences of leaf code $c_j^{(L)}$ given the presence of leaf code $c_i^{(L)}$ within a visit. We then derive the co-occurrence matrices at higher levels by aggregating the co-occurrence counts of their children as follows:
\begin{equation}
  Q^{(l)}_{pq} = \sum_{c_i^{(L)} \in \mathcal{C}(p)} \sum_{c_j^{(L)} \in \mathcal{C}(q)} Q_{ij}, \quad l = 1, \ldots, L
\end{equation}
where \(\mathcal{C}(p) = \mathcal{P}^{L}(c_p^{(l)})\) and \(\mathcal{C}(q)= \mathcal{P}^{L}(c_q^{(l)})\) denote the sets of leaf-level children of parent-level nodes \(c_p^{(l)}\) and \(c_q^{(l)}\), respectively. Next, we derive the co-occurrence conditional probability matrix \(P^{(l)}\) from the count matrix \(Q^{(l)}\) by normalizing each entry with the total occurrences of the corresponding node \(p\), expressed as: \(P^{(l)}_{pq} = Q^{(l)}_{pq} / \sum_{j} Q^{(l)}_{pj}, \quad \text{for } l = 1:L\). The co-occurrence probability matrix is then used to define edges in the graph. Specifically, an edge between nodes \(p\) and \(q\) is included if the co-occurrence probability exceeds a threshold \(\tau^{(l)}\). This binarization generates the adjacency matrix \(\mathcal{A}_h^{(l)}\) from \(P^{(l)}\) as: \(\mathcal{A}_h^{(l)} = \mathbb{I}(P^{(l)} \geq \tau)\), where indicator function \(\mathbb{I}(\cdot)\) returns 1 if true and 0 otherwise. Consequently, we construct the horizontal graphs \(\mathcal{G}_{h}^{(l)} = (\mathcal{V}^{(l)}, \mathcal{E}^{(l)})\), where \(\mathcal{V}^{(l)} = \mathbf{c}^{(l)}\) and \(\mathcal{E}^{(l)} = \mathcal{A}_h^{(l)}\), with \(N^{(l)}\) nodes and \(M^{(l)}\) edges at the \(l\)-th level of the ontology. Note that since \(P^{(l)}_{pq} \neq P^{(l)}_{qp}\) in general, \(\mathcal{A}_h^{(l)}\) is not necessarily a symmetric matrix. Therefore, \(\mathcal{G}_h^{(l)}\) represents a directed graph. We employ a regular graph neural operator, such as GAT \citep{velivckovic2017graph}, leveraging the multihead attention mechanism, to encode medical codes in each level:
\begin{equation}
\label{eq:H-gat}
\mathbf{X}^{(l)}_{(k+1)} = \text{GAT}_{\mathcal{G}_h^{(l)}}\left(\mathbf{X}^{(l)}_{(k)}, \mathbf{\mathcal{A}}_h^{(l)}\right), \quad l = 1, \ldots, L
\end{equation}
where \(\mathbf{X}^{(l)}_{(k+1)} \in \mathbb{R}^{N^{(l)} \times d}\) and \(\mathbf{X}^{(l)}_{(k)} \in \mathbb{R}^{N^{(l)} \times d}\) denote the node features at the \((k+1)\)-th and \(k\)-th layers of \(\mathcal{G}_h^{(l)}\), respectively. Alternatively, for the leaf level of the ontology (\(l=L\)), we can utilize a hypergraph structure due to its robust capability to capture the high-order complex relationships between visits and medical codes \citep{xu2023hypergraph, xu2024ram, cai2022hypergraph}. In this approach, visits are treated as hyperedges \(\mathcal{E}^{(L)} = V\), and leaf-level medical codes are treated as nodes \(\mathcal{V}^{(L)} = \bm{c}^{(L)}\). This allows us to construct \(\mathcal{G}_{h}^{(L)} = (\mathcal{V}^{(L)}, \mathcal{E}^{(L)})\) at the leaf level of the ontology with \(N^{(L)}\) nodes and \(M^{(L)}\) hyperedges. We employ the Hypergraph Attention Network (HAT) \citep{bai2021hypergraph} to encode this leaf-level hypergraph:
\begin{equation}
\label{eq:hat}
\mathbf{X}^{(L)}_{(k+1)} = \text{HAT}_{\mathcal{G}_{h}^{(L)}}\left(\mathbf{X}^{(L)}_{(k)}, \mathbf{\mathcal{H}}^{(L)}_h\right)
\end{equation}
where \(\mathcal{H}^{(L)}_h \in \mathbb{R}^{N^{(L)} \times M^{(L)}}\) is the hypergraph incidence matrix mapping nodes to edges.

We avoid using hypergraphs for the ancestral levels, which involve fewer nodes. Employing hypergraphs at these levels would necessitate defining hyperedges with visits, resulting in an excessive number of hyperedges relative to the fewer nodes. This leads to a dense graph where nodes are redundantly connected, causing embeddings to become overly similar, despite representing distinct entities. Instead, we employ a regular graph structure, allowing for one-to-one edges based on co-occurrence, with edge inclusion controlled by a predefined co-occurrence probability threshold.

\subsection{Step 3: Vertical Message Passing (VMP)}  
\vspace{0.2cm}
Furthermore, we leverage vertical information propagation across the hierarchical levels within the nodes of each individual ontology separately in the Meta-KG. The message passing over ancestor to descendant levels and vice versa enables the information sharing across concepts at different levels of granularity. Inspired by ideas of attention propagation in \citep{zhang2020hierarchical} and the ``two-round propagation'' approach in \citep{pearl2022reverend}, we design the VMP module, a two-round hierarchy-aware graph-based encoding technique that integrates information across all ontology levels.

The first round, the bottom-up propagation, called Hierarchical Graph Information Propagation (HGIP), adaptively updates each concept node in the ontology as a convex combination of itself and its child concepts using multi-head attention mechanism. This process begins by constructing a series of sequential directed vertical subgraphs consisting of a pair of adjacent ontology levels, starting with $\mathcal{G}_v^{(L-1)}$ (connecting level $L$ to $L-1$) and continuing up to $\mathcal{G}_v^{(1)}$ (connecting level 2 to level 1). Edges in each subgraph are defined by parent-child relationships, with directed edges from nodes in level \(l\) to their parent nodes in level \(l-1\), forming the adjacency matrix \(\mathcal{A}_v^{(l)}\) for the vertical subgraph \(\mathcal{G}_v^{(l)}\). We then apply a graph attention operator, such as GAT, to each subgraph, encoding the hierarchical structure in a bottom-up manner. Starting with $\mathcal{G}_v^{(L-1)}$, parent node embeddings in level $L-1$ are updated by aggregating information from their children in level $L$ using multi-head attention. These updated results are then incorporated into $\mathcal{G}_v^{(L-2)}$, where the child embeddings are the updated parent nodes from $\mathcal{G}_v^{(L-1)}$. This sequential process continues to the root level subgraph $\mathcal{G}_v^{(1)}$, propagating information throughout the ontology. Defining step $s = 0, \dots, L-1$, the sequential bottom-up learning process of HGIP is expressed as:
\vspace{-0.4cm}

\begin{equation}
\label{eq:hgip}
[\mathbf{X}^{(L-s)}_{(k+1)}, \mathbf{X}^{(L-s+1)}_{(k+1)}] = \text{GAT}_{\mathcal{G}_v^{(L-s)}}\left([\mathbf{X}^{(L-s)}_{(k)}, \mathbf{X}^{(L-s+1)}_{(k)}], \mathbf{\mathcal{A}}_v^{(L-s)}\right)
\end{equation}
where \(\mathbf{X}^{(l-s)}_{(k)} \in \mathbb{R}^{N^{(l-s)} \times d}\) and \(\mathbf{X}^{(l-s+1)}_{(k)} \in \mathbb{R}^{N^{(l-s+1)} \times d}\) represent the embeddings of parent and child nodes at the \((l-s)\)-th and \((l-s+1)\)-th level of the ontology, which reside in the subgraph \(\mathcal{G}_v^{(l-s)}\). The operator \([,]\) denotes the concatenation of these embeddings, forming the node set of \(\mathcal{G}_v^{(L-s)}\).

Unlike the previous approaches defining a single graph for vertical message passing, we represent an ontology as a series of sequential subgraphs, where each corresponds to a pair of adjacent levels. This approach, while ensuring that each parent node in the hierarchy contains the curated distilled information of all its descendants in lower levels, respects the order of node embedding updates across the main graph, incorporating the hierarchical order. This enhances the bottom-up HAP \citep{zhang2020hierarchical} in two ways: 1) it employs a sequential GNN structure for efficient, parallel node updates, and 2) it integrates a multi-head attention mechanism to compute attention weights, enabling expressive multi-view representations and addressing the potential inconsistencies between EHR co-occurrences and ontologies \citep{song2019medical}.


Second round, we apply GRAM \citep{choi2017gram} to compute the final representation of the leaf-level nodes by adaptively aggregating information from their ancestors using attention mechanism. The final representation \(z_{i} \in \mathbb{R}^{d_{c}}\) of each leaf-level code \(c_i^{(L)}\), where \(i = 1:N^{(L)}\), is computed as a convex combination of child embedding \(x_i^{(L)}\) and all its ancestors' embeddings:
\begin{equation}
\label{eq:gram}
\text{GRAM}: \quad z_{i} = \sum_{l=1}^{L} \alpha_{il} \mathcal{P}^{l}(x_{i}^{(L)}), \quad \quad \alpha_{il} \geq 0, \quad \text{for} \quad l = 1, \ldots, L
\end{equation}
where \(\alpha_{il} \in \mathbb{R}^{+}\) denotes the attention weight for the code embedding \(\mathcal{P}^{l}(x_{i}^{(L)})\) in computing \(z_{i}\). The attention weight \(\alpha_{il}\) is computed using the softmax function as:
\begin{equation}
\alpha_{il} = \frac{\exp(f(x^{(L)}_{i}, \mathcal{P}^{l}(x_{i}^{(L)}))}{\sum_{k=1}^{L} \exp(f(x^{(L)}_{i}, \mathcal{P}^{k}(x_{i}^{(L)}))}
\end{equation}
where \(f(a,b)\) is an MLP producing a scalar energy for the raw attention between \(a\) and \(b\), which the softmax normalizes into values between 0 and 1.

\subsection{Integrating \model{} into EHR Models}

We introduce \model{} as a medical concept encoder designed to be connected to different EHR predictive models and trained end-to-end, enhancing concept representation learning and improving predictive performance. The final medical concept (EHR code) embeddings produced by \model{}, \(\mathbf{z}_{1}, \mathbf{z}_{2}, \dots, \mathbf{z}_{N^{(L)}}\) form the embedding matrix \(\mathbf{Z} \in \mathbb{R}^{d \times N^{(L)}}\), where \(\mathbf{z}_{i}\) is the \(i\)-th column of \(\mathbf{Z}\). This embedding matrix is then used in a downstream task. For instance, for sequential diagnosis prediction, which maps a sequence of visits to the predicted diagnoses of the next visit, we have \(f: \{ V_{1}, V_{2}, \ldots, V_{t}\} \rightarrow \mathbf{\hat{y}}_{t+1}\), where \(\mathbf{\hat{y}}_{t+1} \in \mathbb{R}^{N^{(L)}_{dx}}\) is a multi-hot vector, with \(N^{(L)}_{dx}\) denoting the total number of diagnosis codes:
\begin{equation}
\label{eq:downstream_task}
\begin{aligned}
    \mathbf{Z} = [\mathbf{z}_{1}, \mathbf{z}_{2}, \ldots, \mathbf{z}_{N^{(L)}}] &\gets \text{\model{}}(\mathbf{x}^{(L)}_{1},\mathbf{x}^{(L)}_{2}, \ldots, \mathbf{x}^{(L)}_{N^{(L)}}) \\
    \mathbf{v}_{1}, \mathbf{v}_{2}, \ldots, \mathbf{v}_{t} &= \mathbf{Z}[\mathbf{u}_{1}, \mathbf{u}_{2}, \ldots, \mathbf{u}_{t}] \\
    \mathbf{h}_{p} &= \text{Model}(\mathbf{v}_{1}, \mathbf{v}_{2}, \ldots, \mathbf{v}_{t}) \\
    \mathbf{\hat{y}}_{t+1} &= \text{Sigmoid}(\mathbf{W} \mathbf{h}_{p} + \mathbf{b})
\end{aligned}
\end{equation}
For each visit \(V_{t}\), we obtain a representation \(\mathbf{v}_t \in \mathbb{R}^{d}\) by multiplying the final embedding matrix \(\mathbf{Z}\) with a multi-hot vector \(\mathbf{u}_t =\{0,1\}^{N^{(L)}}\), which represents clinical events existence in the visit. The sequence of visit representations \(\{ \mathbf{v}_{1}, \mathbf{v}_{2}, \ldots, \mathbf{v}_{t}\}\) serves as input to a main health model, \(\text{Model}(\cdot)\) (e.g., Retain \cite{choi2016retain}), producing the patient representation \(\mathbf{h}_{p} \in \mathbb{R}^{d}\).
The final prediction is computed by applying a Sigmoid function to the linear transformation of \(\mathbf{h}_{p}\), with \(\mathbf{W} \in \mathbb{R}^{N^{(L)} \times d}\) and \(\mathbf{b} \in \mathbb{R}^{N^{(L)}}\) as the weight and bias, respectively. The output \(\mathbf{\hat{y}}_{t+1}\) is the predicted vector in \(\mathbb{R}^{N^{(L)}}\). The loss at each timestamp is the cross-entropy between the ground truth \(y_{t+1}\) and prediction \(\hat{y}_{t+1}\).

\begin{table}[h]
\centering
\addtolength{\tabcolsep}{-0.4mm}
\vspace{-3mm}
\caption{Data statistics for MIMIC-III and MIMIC-IV.}
\vspace{-3mm}
\label{tab:Data-Statistics}
\small
\renewcommand{\arraystretch}{0.95}
\makebox[\linewidth]{
    \begin{tabular}{@{}lcc@{}}
    \hline
    \textbf{Metric} & \textbf{MIMIC-III} & \textbf{MIMIC-IV} \\
    \hline
    \# Patients              & 7,515   & 18,829  \\
    \# Visits (samples)                & 12,430  & 25,028  \\
    \# Labels/sample         & 13.32   & 11.89   \\
    \# Unique conditions (ICD) & 4,283  & 7,054   \\
    \# Conditions/sample        & 29.02   & 66.84   \\
    \# Drugs/sample             & 70.10   & 118.17  \\
    \# Unique drugs             & 471     & 510     \\
    \# Procedures/sample        & 7.01    & 5.77    \\
    \# Unique procedures        & 1,328   & 2,033   \\
    \hline
    \end{tabular}
}
\vspace{-4mm}
\end{table}
\vspace{-2mm}
\section{Experimental Setting}

\begin{figure*}[t]
    \setlength{\abovecaptionskip}{3pt} 
    \begin{center}
    \includegraphics[width=1.0\linewidth]{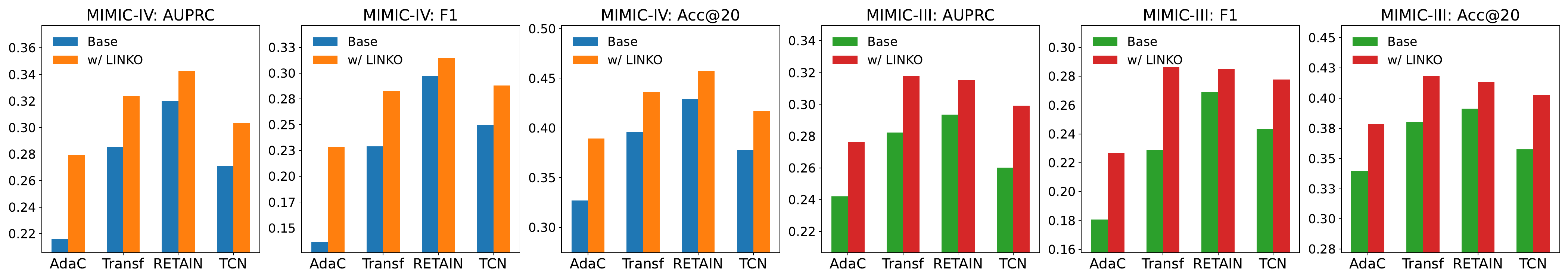}
    \caption{Performance enhancement evaluation before and after integrating \model{} into four diagnosis prediction models (plug-in analysis), using the MIMIC-III and MIMIC-IV datasets.} 
    \label{fig:enhancement}
    \end{center}
\end{figure*}

\begin{table*}[ht]
    \centering
    \footnotesize
    \caption{Performance comparison and ablation study on MIMIC-III and MIMIC-IV. General Performance includes AUPRC, F1 score, and Acc@k. Diagnosis labels are grouped into four frequency bands (0–25\%, 25–50\%, 50–75\%, 75–100\%), and corresponding AUPRC scores are reported under Label Category Performance. The first row shows the base Transformer model, while subsequent rows represent variants with added medical concept encoders (e.g., GRAM = Base + GRAM). Reported values are means with 95\% confidence intervals.}
    \vspace{-2mm}
    \renewcommand{\arraystretch}{1.05}
    \setlength{\tabcolsep}{6pt}
    \begin{tabular}{c|l ccccc cccc}
        \hline
        \textbf{} & \multirow{2}{*}{\textbf{Model}} & \multicolumn{5}{c }{\textbf{General Performance}} & \multicolumn{4}{c}{\textbf{Label Category Performance (AUPRC)}} \\
        \cline{3-7} \cline{8-11}
         & & \textbf{AUPRC} & \textbf{F1} & \textbf{Acc@15} & \textbf{Acc@20} & \textbf{Acc@30} & \textbf{0-25\%} & \textbf{25-50\%} & \textbf{50-75\%} & \textbf{75-100\%} \\
        \hline
        \multirow{8}{*}{\rotatebox{90}{\textbf{MIMIC-IV}}}
        & Base  & $28.54_{\pm0.39}$ & $22.90_{\pm0.10}$ & $37.88_{\pm0.53}$ & $39.53_{\pm0.60}$ & $44.10_{\pm0.52}$ & $28.33_{\pm0.44}$ & $58.73_{\pm0.86}$ & $46.54_{\pm0.30}$ & $67.55_{\pm0.18}$ \\
        & GRAM  & $29.96_{\pm0.45}$ & $24.17_{\pm0.18}$ & $39.42_{\pm0.25}$ & $41.23_{\pm0.48}$ & $45.90_{\pm0.61}$ & $29.73_{\pm0.53}$ & $60.62_{\pm0.27}$ & $47.67_{\pm0.27}$ & $68.45_{\pm0.21}$ \\
        & MMORE  & $30.06_{\pm0.25}$ & $25.11_{\pm1.60}$ & $39.56_{\pm0.18}$ & $41.46_{\pm0.32}$ & $46.19_{\pm0.34}$ & $29.84_{\pm0.22}$ & $60.60_{\pm0.37}$ & $48.84_{\pm0.34}$ & $68.07_{\pm1.3}$ \\
        & KAME & $29.13_{\pm0.32}$ & $23.39_{\pm0.32}$ & $39.18_{\pm0.32}$ & $40.28_{\pm0.32}$ & $45.67_{\pm0.25}$ & $28.84_{\pm0.32}$ & $59.62_{\pm0.32}$ & $47.21_{\pm0.42}$ & $68.08_{\pm0.32}$ \\
        & G-BERT  & $30.13_{\pm0.36}$ & $25.19_{\pm0.33}$ & $39.65_{\pm0.73}$ & $41.66_{\pm0.81}$ & $46.29_{\pm0.95}$ & $29.91_{\pm0.59}$ & $60.83_{\pm0.69}$ & $49.88_{\pm0.74}$ & $69.03_{\pm1.0}$ \\
        & HAP  & $30.02_{\pm0.23}$ & $24.22_{\pm1.30}$ & $39.55_{\pm0.23}$ & $41.37_{\pm0.37}$ & $46.22_{\pm0.33}$ & $29.64_{\pm0.34}$ & $60.76_{\pm0.35}$ & $49.53_{\pm0.33}$ & $68.94_{\pm1.4}$ \\
        & ADORE & $30.11_{\pm0.34}$ & $25.21_{\pm0.87}$ & $39.87_{\pm0.24}$ & $41.51_{\pm0.42}$ & $46.25_{\pm0.34}$ & $30.15_{\pm0.30}$ & $60.89_{\pm0.45}$ & $49.44_{\pm1.02}$ & $68.18_{\pm0.93}$ \\
        & KAMPNet & $30.21_{\pm0.23}$ & $25.32_{\pm1.10}$ & $40.18_{\pm0.42}$ & $41.88_{\pm0.54}$ & $46.41_{\pm0.56}$ & $30.21_{\pm0.34}$ & $61.05_{\pm0.89}$ & $48.93_{\pm0.53}$ & $70.30_{\pm1.0}$ \\
        & $\text{\model{}}_{\text{w/ GAT}}$ & $32.13_{\pm0.12}$ & $28.56_{\pm0.09}$ & $\mathbf{42.60_{\pm0.14}}$ & $43.56_{\pm0.14}$ & $48.12_{\pm0.18}$ & $31.83_{\pm0.13}$ & $63.03_{\pm0.26}$ & $50.96_{\pm0.40}$ & $71.35_{\pm0.29}$ \\
        & $\text{\model{}}_{\text{w/ HAT}}$ & $\mathbf{32.38}_{\pm\mathbf{0.32}}$ & $\mathbf{30.02_{\pm0.13}}$ & $42.05_{\pm0.65}$ & $\mathbf{43.70}_{\pm\mathbf{0.66}}$ & $\mathbf{48.63}_{\pm\mathbf{0.73}}$ & $\mathbf{32.31}_{\pm\mathbf{0.41}}$ & $\mathbf{63.25}_{\pm\mathbf{0.73}}$ & $\mathbf{51.17}_{\pm\mathbf{0.23}}$ & $\mathbf{71.58}_{\pm\mathbf{0.12}}$ \\
        & $\text{\hspace{1cm} w/o HGIP}$ & $30.65_{\pm0.28}$ & $26.19_{\pm0.22}$ & $40.16_{\pm0.20}$ & $42.03_{\pm0.21}$ & $46.81_{\pm0.40}$ & $30.36_{\pm0.45}$ & $61.34_{\pm0.13}$ & $49.15_{\pm0.19}$ & $68.63_{\pm0.55}$ \\
        & $\text{\hspace{1cm} w/o LLM}$ & $30.86_{\pm0.59}$ & $26.69_{\pm0.52}$ & $40.40_{\pm0.83}$ & $42.06_{\pm0.75}$ & $46.54_{\pm0.88}$ & $30.66_{\pm0.68}$ & $61.42_{\pm0.10}$ & $48.35_{\pm0.18}$ & $67.10_{\pm0.32}$ \\
        & $\text{\hspace{1cm} w/o leaf-HMP}$ & $30.70_{\pm0.33}$ & $26.89_{\pm0.88}$ & $40.27_{\pm0.43}$ & $42.09_{\pm0.38}$ & $46.76_{\pm0.60}$ & $30.41_{\pm0.45}$ & $61.30_{\pm0.99}$ & $48.77_{\pm0.79}$ & $69.46_{\pm0.99}$ \\
        & $\text{\hspace{1cm} w/o parent-HMP}$ & $30.32_{\pm0.69}$ & $25.49_{\pm0.25}$ & $39.83_{\pm0.59}$ & $41.72_{\pm0.92}$ & $46.47_{\pm0.90}$ & $30.07_{\pm0.78}$ & $61.40_{\pm0.72}$ & $46.46_{\pm0.23}$ & $68.06_{\pm0.29}$ \\
        & $\text{\hspace{1cm} w/o HMP}$ & $30.10_{\pm0.39}$ & $25.26_{\pm0.88}$ & $39.67_{\pm0.41}$ & $41.50_{\pm0.34}$ & $46.25_{\pm0.53}$ & $29.85_{\pm0.51}$ & $60.94_{\pm0.75}$ & $47.39_{\pm0.28}$ & $67.33_{\pm0.62}$ \\
        \hline
        \multirow{8}{*}{\rotatebox{90}{\textbf{MIMIC-III}}} 
        & Base & $28.23_{\pm0.24}$ & $22.36_{\pm0.33}$ & $36.15_{\pm0.22}$ & $38.03_{\pm0.34}$ & $43.19_{\pm0.43}$ & $28.07_{\pm0.28}$ & $54.28_{\pm0.10}$ & $49.89_{\pm0.29}$ & $73.34_{\pm0.26}$ \\
        & GRAM & $28.99_{\pm0.34}$ & $24.10_{\pm0.48}$ & $37.17_{\pm0.36}$ & $39.13_{\pm0.52}$ & $44.32_{\pm0.51}$ & $28.79_{\pm0.41}$ & $54.81_{\pm1.3}$ & $50.58_{\pm0.34}$ & $74.44_{\pm0.77}$ \\
        & MMORE & $29.11_{\pm0.38}$ & $23.67_{\pm0.70}$ & $37.15_{\pm0.46}$ & $39.14_{\pm0.53}$ & $44.36_{\pm0.50}$ & $28.87_{\pm0.43}$ & $54.92_{\pm0.87}$ & $51.29_{\pm0.25}$ & $74.42_{\pm0.30}$ \\
        & KAME & $28.52_{\pm0.32}$ & $23.18_{\pm0.09}$ & $36.76_{\pm0.47}$ & $38.36_{\pm0.45}$ & $43.75_{\pm0.36}$ & $28.19_{\pm0.38}$ & $55.13_{\pm0.16}$ & $50.09_{\pm0.32}$ & $73.79_{\pm0.23}$ \\
        & G-BERT & $29.22_{\pm0.81}$ & $24.92_{\pm0.83}$ & $37.36_{\pm0.12}$ & $39.55_{\pm0.12}$ & $45.15_{\pm0.44}$ & $29.16_{\pm0.10}$ & $54.47_{\pm0.99}$ & $52.45_{\pm0.62}$ & $75.40_{\pm0.18}$ \\
        & HAP & $29.28_{\pm0.41}$ & $23.25_{\pm0.88}$ & $37.47_{\pm0.56}$ & $39.64_{\pm0.55}$ & $45.05_{\pm0.57}$ & $29.10_{\pm0.45}$ & $55.11_{\pm0.99}$ & $52.19_{\pm0.27}$ & $75.50_{\pm0.24}$ \\
        & ADORE & $29.31_{\pm0.29}$ & $23.61_{\pm0.81}$ & $37.53_{\pm0.31}$ & $39.68_{\pm0.32}$ & $45.01_{\pm0.58}$ & $29.18_{\pm0.73}$ & $55.14_{\pm0.42}$ & $52.14_{\pm0.45}$ & $75.41_{\pm0.27}$ \\
        & KAMPNet & $29.38_{\pm0.65}$ & $25.01_{\pm0.72}$ & $37.83_{\pm0.39}$ & $39.91_{\pm0.61}$ & $45.26_{\pm0.53}$ & $29.31_{\pm0.45}$ & $55.20_{\pm0.87}$ & $52.56_{\pm0.34}$ & $75.91_{\pm0.29}$ \\
        & $\text{\model{}}_{\text{w/ HAT}}$ & $30.78_{\pm0.99}$ & $27.67_{\pm0.23}$ & $39.11_{\pm0.92}$ & $41.18_{\pm0.73}$ & $46.65_{\pm0.83}$ & $30.72_{\pm0.85}$ & $54.74_{\pm0.19}$ & $\mathbf{57.16}_{\pm\mathbf{0.26}}$ & $79.23_{\pm0.10}$ \\
        & $\text{\model{}}_{\text{w/ GAT}}$ & $\mathbf{31.79}_{\pm\mathbf{0.11}}$ & $\mathbf{28.66_{\pm0.18}}$ & $\mathbf{39.95}_{\pm\mathbf{0.11}}$ & $\mathbf{41.84}_{\pm\mathbf{0.14}}$ & $\mathbf{46.96}_{\pm\mathbf{0.96}}$ & $\mathbf{31.62}_{\pm\mathbf{1.09}}$ & $\mathbf{55.68_{\pm0.15}}$ & $56.90_{\pm0.59}$ & $\mathbf{80.76}_{\pm\mathbf{0.47}}$ \\
        &$\text{\hspace{1cm} w/o HGIP}$ & $30.25_{\pm0.37}$ & $25.75_{\pm0.14}$ & $38.54_{\pm0.51}$ & $40.47_{\pm0.55}$ & $46.00_{\pm0.66}$ & $30.04_{\pm0.44}$ & $55.61_{\pm0.12}$ & $53.36_{\pm0.31}$ & $77.00_{\pm0.22}$ \\
        & $\text{\hspace{1cm} w/o LLM}$ & $30.38_{\pm0.10}$ & $25.88_{\pm0.90}$ & $38.23_{\pm0.57}$ & $40.38_{\pm0.66}$ & $45.82_{\pm0.90}$ & $30.33_{\pm0.10}$ & $55.05_{\pm0.99}$ & $55.03_{\pm0.64}$ & $79.16_{\pm0.33}$ \\
        & $\text{\hspace{1cm} w/o leaf-HMP}$ & $30.09_{\pm0.43}$ & $25.44_{\pm0.15}$ & $38.25_{\pm0.54}$ & $40.35_{\pm0.54}$ & $45.92_{\pm0.59}$ & $29.88_{\pm0.51}$ & $55.57_{\pm0.57}$ & $53.62_{\pm0.30}$ & $76.68_{\pm0.17}$ \\
        & $\text{\hspace{1cm} w/o parent-HMP}$ & $29.76_{\pm0.43}$ & $26.02_{\pm0.10}$ & $37.97_{\pm0.39}$ & $39.93_{\pm0.56}$ & $45.43_{\pm0.52}$ & $29.53_{\pm0.48}$ & $55.24_{\pm0.88}$ & $52.48_{\pm0.26}$ & $75.78_{\pm0.17}$ \\
        & $\text{\hspace{1cm} w/o HMP}$ & $29.41_{\pm0.40}$ & $25.28_{\pm0.88}$ & $37.55_{\pm0.41}$ & $39.61_{\pm0.52}$ & $45.25_{\pm0.60}$ & $29.26_{\pm0.50}$ & $54.52_{\pm0.63}$ & $52.58_{\pm0.20}$ & $75.46_{\pm0.21}$ \\ 
        \hline
    \end{tabular}
    \label{tab:baseline_comparison_w_rare_codes}
\end{table*}

\textbf{Datasets: }We utilize two publicly available datasets: MIMIC-III \citep{johnson2016mimic} and MIMIC-IV \citep{johnson2023mimic}. MIMIC-III (2001–2012) uses ICD-9 codes, while MIMIC-IV (2008–2019) includes both ICD-9 and ICD-10 and provides more comprehensive longitudinal data. Prescription codes in both datasets follow the National Drug Code (NDC) system, which we map to the Anatomical Therapeutic Chemical (ATC) Classification. Table~\ref{tab:Data-Statistics} presents cohort statistics. The task is a multi-label sequential prediction involving identifying the ICD-9 diagnosis codes for the next visit from a highly imbalanced and vast label space, including 4,283 unique codes in MIMIC-III and 8,818 in MIMIC-IV.

\noindent\textbf{Implementations: }We report the mean and confidence intervals of the results after 5-fold experimentation. \model{} uses 4 attention heads for horizontal graphs, 2 for vertical graphs, dropout rates of 0.1 and 0.2, respectively, and a shared embedding dimension of \(d = 256\) for all nodes in the Meta-KG. We use a 3-level hierarchy (\(L = 3\)) for the ICD-9 diagnosis, ICD-9 procedure, and ATC drug ontologies in our experiments.

\noindent\textbf{Evaluation Metrics:} 
\textbf{(1) AUPRC}: Measures the area under the precision-recall curve, capturing the trade-off between precision and recall across different thresholds. We also report AUPRC scores stratified by label frequency to assess performance across code rarity levels.
\textbf{(2) Acc@k}: The number of correct diagnosis codes among the top \(k\) predictions, divided by \(\min(k, \|y_t\|)\), where \(\|y_t\|\) is the number of ground-truth labels in the \((t+1)\)-th visit. 
\textbf{(3) F1-score}: The harmonic mean of precision and recall, providing a balanced measure of model performance.

\vspace{-0.8cm}
\section{Evaluation Results}

To evaluate our proposed model, we investigate the following research questions:  
\textbf{RQ1:} Does our method enhance the performance of EHR models as a complementary medical concept encoder?  
\textbf{RQ2:} How does \model{} perform compared to existing medical code encoders?  
\textbf{RQ3:} What is the impact of each component of \model{} on performance?  
\textbf{RQ4:} How do the prompting strategy and its included information affect performance?  
\textbf{RQ5:} How does our method mitigate data insufficiency limitations and rare disease predictions?  
\textbf{RQ6:} How does \model{} learn representations for individual codes?  


\begin{table*}[t]
    \centering
    \small
    \caption{Performance analysis of different prompting strategies on MIMIC-III and MIMIC-IV. The reported values include means and 95\% confidence intervals.}
    \vspace{-2mm}
    \renewcommand{\arraystretch}{1.05}
    \setlength{\tabcolsep}{3pt}
    \begin{tabular}{l|ccccc|ccccc}
        \hline
        \multirow{2}{*}{\textbf{Prompt Information}} & \multicolumn{5}{c|}{\textbf{MIMIC-IV}} & \multicolumn{5}{c}{\textbf{MIMIC-III}} \\
        \cline{2-11}
        & \textbf{AUPRC} & \textbf{F1} & \textbf{Acc@15} & \textbf{Acc@20} & \textbf{Acc@30} 
        & \textbf{AUPRC} & \textbf{F1} & \textbf{Acc@15} & \textbf{Acc@20} & \textbf{Acc@30} \\
        \hline        
        no LLM & $30.86_{\pm0.59}$ & $26.69_{\pm0.52}$ & $40.40_{\pm0.83}$ & $42.06_{\pm0.75}$ & $46.54_{\pm0.88}$ 
        & $30.38_{\pm0.10}$ & $25.88_{\pm0.90}$ & $38.23_{\pm0.57}$ & $40.38_{\pm0.66}$ & $45.82_{\pm0.90}$ \\
        
        type-code-noise & $31.28_{\pm0.10}$ & $28.46_{\pm0.30}$ & $40.99_{\pm0.75}$ & $43.25_{\pm0.10}$ & $47.39_{\pm0.11}$ 
        & $30.57_{\pm0.74}$ & $27.88_{\pm0.21}$ & $38.99_{\pm0.46}$ & $40.96_{\pm0.57}$ & $46.57_{\pm0.63}$ \\
        
        type-code & $31.96_{\pm0.40}$ & $27.77_{\pm0.67}$ & $41.57_{\pm0.35}$ & $43.24_{\pm0.42}$ & $47.92_{\pm0.50}$ 
        & $31.15_{\pm0.98}$ & $27.89_{\pm0.10}$ & $39.28_{\pm0.69}$ & $41.26_{\pm0.44}$ & $46.79_{\pm0.53}$ \\
        
        type-code-concept & $32.00_{\pm0.25}$ & $29.07_{\pm0.56}$ & $41.73_{\pm0.74}$ & $43.52_{\pm0.83}$ & $48.17_{\pm0.78}$ 
        & $31.18_{\pm1.00}$ & $28.35_{\pm0.65}$ & $39.39_{\pm0.91}$ & $41.55_{\pm0.83}$ & $46.90_{\pm0.94}$ \\
        
        type-code-concept-children & $32.05_{\pm1.00}$ & $29.03_{\pm1.02}$ & $41.57_{\pm0.11}$ & $43.41_{\pm0.13}$ & $47.94_{\pm0.14}$ 
        & $31.09_{\pm1.00}$ & $27.67_{\pm1.02}$ & $39.23_{\pm0.13}$ & $41.18_{\pm0.10}$ & $46.66_{\pm0.99}$ \\
        
        type-code-concept-parent & $32.35_{\pm0.50}$ & $28.27_{\pm0.48}$ & $42.01_{\pm0.73}$ & $43.59_{\pm0.38}$ & $48.49_{\pm0.41}$ 
        & $31.21_{\pm0.95}$ & $28.48_{\pm0.29}$ & $39.46_{\pm0.79}$ & $41.64_{\pm0.10}$ & $46.91_{\pm0.97}$ \\
        
        type-code-concept-parent-task & $\mathbf{32.38}_{\pm\mathbf{0.32}}$ & $\mathbf{30.02_{\pm0.13}}$ & $\mathbf{42.05}_{\pm\mathbf{0.65}}$ & $\mathbf{43.70}_{\pm\mathbf{0.66}}$ & $\mathbf{48.63}_{\pm\mathbf{0.73}}$ 
        & $\mathbf{31.79}_{\pm\mathbf{0.11}}$ & $\mathbf{28.66_{\pm0.18}}$ & $\mathbf{39.95}_{\pm\mathbf{0.11}}$ & $\mathbf{41.84}_{\pm\mathbf{0.14}}$ & $\mathbf{46.96}_{\pm\mathbf{0.96}}$ \\
        
        $\text{\hspace{1cm} w/ LLM-emb-freezed}$ & $22.13_{\pm0.77}$ & $13.35_{\pm0.89}$ & $30.79_{\pm0.90}$ & $32.91_{\pm0.92}$ & $37.70_{\pm0.97}$ 
        & $25.75_{\pm0.37}$ & $19.96_{\pm0.36}$ & $33.82_{\pm0.39}$ & $36.28_{\pm0.38}$ & $42.00_{\pm0.38}$ \\
        \hline
    \end{tabular}
    \label{tab:llm_prompting_analysis}
\end{table*}

\begin{table*}[t]
    \centering
    \small
    \caption{Performance analysis of training on different combinations of medical concept types with/without Multi-level integration of ontologies on MIMIC-III and MIMIC-IV. The reported values include means and 95\% confidence intervals.}
    \vspace{-2mm}
    \renewcommand{\arraystretch}{1.05}
    \setlength{\tabcolsep}{4pt}  
    \begin{tabular}{c|l|ccccc|ccccc}
        \hline
        \textbf{} & \multirow{2}{*}{\textbf{Concept Type}} & \multicolumn{5}{c|}{\textbf{w/ Multi-level Integration}} & \multicolumn{5}{c}{\textbf{w/o Multi-level Integration}} \\
        \cline{3-12}
        &  & \textbf{AUPRC} & \textbf{F1} & \textbf{Acc@15} & \textbf{Acc@20} & \textbf{Acc@30} 
        & \textbf{AUPRC} & \textbf{F1} & \textbf{Acc@15} & \textbf{Acc@20} & \textbf{Acc@30} \\
        \hline        

        \multirow{4}{*}{\rotatebox{90}{\textbf{MIMIC-IV}}} 
        & $\text{rx,px}$ & $22.74_{\pm0.04}$ & $16.92_{\pm0.17}$ & $31.56_{\pm0.20}$ & $33.57_{\pm0.27}$ & $38.75_{\pm0.26}$ 
        & $21.35_{\pm0.27}$ & $15.15_{\pm0.26}$ & $30.03_{\pm0.53}$ & $32.21_{\pm0.65}$ & $37.23_{\pm0.39}$ 
         \\

        & $\text{dx,px}$ & $30.67_{\pm0.44}$ & $25.87_{\pm0.13}$ & $40.25_{\pm0.62}$ & $42.21_{\pm0.59}$ & $46.99_{\pm0.42}$ 
        & $29.28_{\pm0.31}$ & $22.49_{\pm0.15}$ & $38.62_{\pm0.31}$ & $40.68_{\pm0.37}$ & $45.56_{\pm0.24}$ \\

        & $\text{dx,rx}$ & $30.86_{\pm0.10}$ & $25.34_{\pm0.23}$ & $40.44_{\pm0.10}$ & $42.53_{\pm0.12}$ & $47.31_{\pm0.10}$ 
         & $30.17_{\pm0.10}$ & $25.08_{\pm0.14}$ & $39.77_{\pm0.57}$ & $41.71_{\pm0.70}$ & $46.59_{\pm0.64}$\\
        
        & $\text{dx,rx,px}$ & $32.38_{\pm0.32}$ & $30.02_{\pm0.13}$ & $42.05_{\pm0.65}$ & $43.70_{\pm0.66}$ & $48.63_{\pm0.73}$ 
       & $30.10_{\pm0.39}$ & $25.26_{\pm0.23}$ & $39.67_{\pm0.41}$ & $41.50_{\pm0.34}$ & $46.25_{\pm0.53}$ \\
       
       \hline

        \multirow{4}{*}{\rotatebox{90}{\textbf{MIMIC-III}}} 
        & $\text{rx,px}$ & $24.24_{\pm0.97}$ & $18.29_{\pm0.97}$ & $31.92_{\pm0.86}$ & $34.43_{\pm0.14}$ & $39.87_{\pm0.13}$ 
        & $22.98_{\pm0.22}$ & $16.55_{\pm0.14}$ & $30.81_{\pm0.31}$ & $32.95_{\pm0.49}$ & $38.61_{\pm0.33}$ \\

        & $\text{dx,px}$ & $30.19_{\pm0.11}$ & $26.40_{\pm0.26}$ & $38.03_{\pm0.81}$ & $40.18_{\pm0.79}$ & $46.03_{\pm0.11}$ 
        & $29.01_{\pm0.67}$ & $24.48_{\pm0.10}$ & $37.04_{\pm0.74}$ & $39.16_{\pm0.61}$ & $44.89_{\pm0.68}$ \\

        & $\text{dx,rx}$ & $30.64_{\pm0.40}$ & $27.06_{\pm0.15}$ & $39.31_{\pm0.58}$ & $41.27_{\pm0.64}$ & $46.62_{\pm0.57}$
        & $29.56_{\pm0.38}$ & $24.44_{\pm0.65}$ & $37.88_{\pm0.61}$ & $39.98_{\pm0.65}$ & $45.46_{\pm0.60}$ \\
        
        & $\text{dx,rx,px}$ 
        & $31.79_{\pm0.11}$ & $28.66_{\pm0.18}$ & $39.95_{\pm0.11}$ & $41.84_{\pm0.14}$ & $46.96_{\pm0.96}$  
        & $29.41_{\pm0.40}$ & $25.28_{\pm0.88}$ & $37.55_{\pm0.41}$ & $39.61_{\pm0.52}$ & $45.25_{\pm0.60}$ \\
        \hline
    \end{tabular}
    \label{tab:onto_colab}
\end{table*}

\vspace{-3mm}
\subsection{RQ1: Plug-in Enhancement Evaluation}
We propose that incorporating our medical concept encoder, \model{}, into existing EHR models boosts downstream performance through concept representation enhancement. To verify this, we integrated \model{} into four diverse EHR models: {(1) \textbf{AdaCare}} \cite{ma2020adacare}, an explainable model modeling biomarker variations to represent health status across time scales, {(2) \textbf{Transformer}} \citep{vaswani2017attention}, which leverages the power of the self-attention mechanism; {(3) \textbf{RETAIN}} \citep{choi2016retain}, an RNN-based model for EHRs that utilizes a two-level reverse time attention mechanism; and {(4) \textbf{TCN}} \citep{bai2018empirical}, which uses causal convolutions to capture temporal dependencies in sequential data. We conducted experiments with each model both with and without \model{}. Figure \ref{fig:enhancement} illustrates that \model{} consistently improves predictive accuracy when integrated with any of the models, validating its effectiveness in concept representation learning.

\vspace{-5mm}
\subsection{RQ2: Baseline Comparison}

We compare our method with several existing ontology-based medical concept encoders: \textbf{GRAM} \citep{choi2017gram}, \textbf{MMORE} \citep{song2019medical}, \textbf{KAME} \citep{ma2018kame}, \textbf{HAP} \citep{zhang2020hierarchical}, \textbf{G-BERT} \citep{shang2019pre}, \textbf{ADORE} \citep{cheong2023adaptive}, and \textbf{KAMPNet} \citep{an2023kampnet}. These encoders are designed for ontology-based augmentation of medical code representation and are added to predictive models as an extension to boost performance. We used the Transformer \cite{vaswani2017attention} as the base diagnosis prediction model. We compared the following setups: (1) the base model with no medical concept encoder; (2) the base model with each of the four existing encoders; and (3) the base model with \model{}. The general performance section in Table \ref{tab:baseline_comparison_w_rare_codes} (right side) shows that \model{}  outperforms the existing encoders in enhancing the predictive performance of the base model, demonstrating its effectiveness as a complementary concept encoder. We also test two different graph encoding techniques for \(\mathcal{G}_h^{(L)}\) in \model{}. Both techniques outperformed baselines: HAT achieved the best performance on MIMIC-IV (8,818 unique dx codes) by capturing higher-order dependencies in its large, sparse code space. In contrast, GAT excelled on MIMIC-III (4,283 unique dx codes), where denser pairwise co-occurrence enabled more effective neighborhood attention. 

To assess the model’s ability to represent and predict rare concepts, diagnosis codes are divided into four frequency bands based on their occurrence: 0–25\%, 25–50\%, 50–75\%, and 75–100\%, with the 0–25\% band indicating the rarest codes. This stratification highlights \model{}’s robustness under data scarcity, as it effectively propagates information to improve rare code representation. As shown in Table~\ref{tab:baseline_comparison_w_rare_codes} (Label Category Performance), \model{} consistently outperforms baselines, especially for rarer code groups.

\vspace{-3mm}
\subsection{RQ3: Ablation Study}

\begin{figure*}[t]
    \setlength{\abovecaptionskip}{5pt} 
    \begin{center}
    \includegraphics[width=0.9\linewidth]{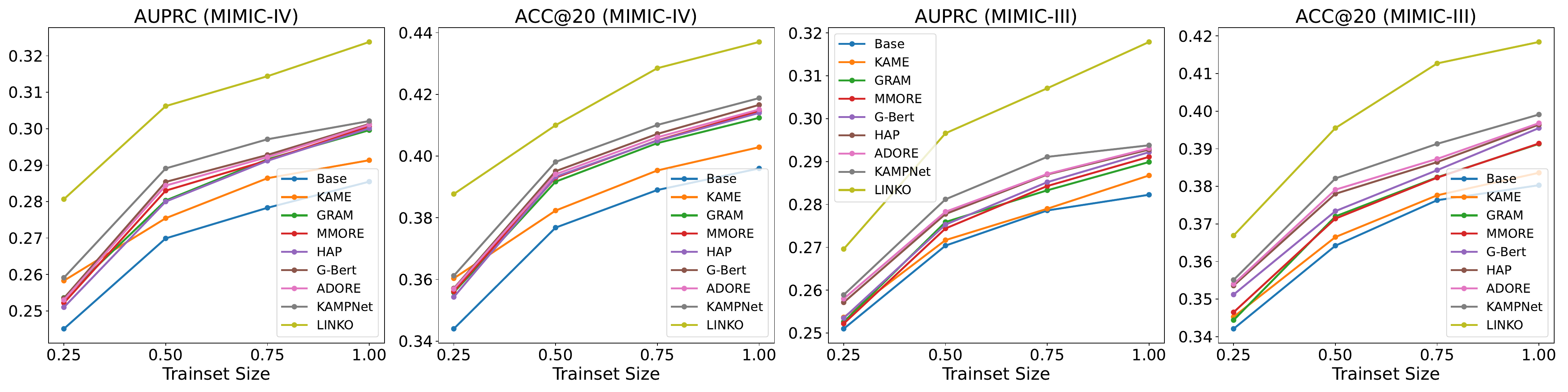}
    \caption{Performance comparison of integration of ontology encoder baselines into the base model (Transformer) across different training set sizes using the MIMIC-III and MIMIC-IV datasets.}
    \label{fig:basline_trainset_size}
    \end{center}
    \vspace{-5mm}
\end{figure*}

\begin{figure*}[h]
\begin{center}
\includegraphics[width=0.9\linewidth]{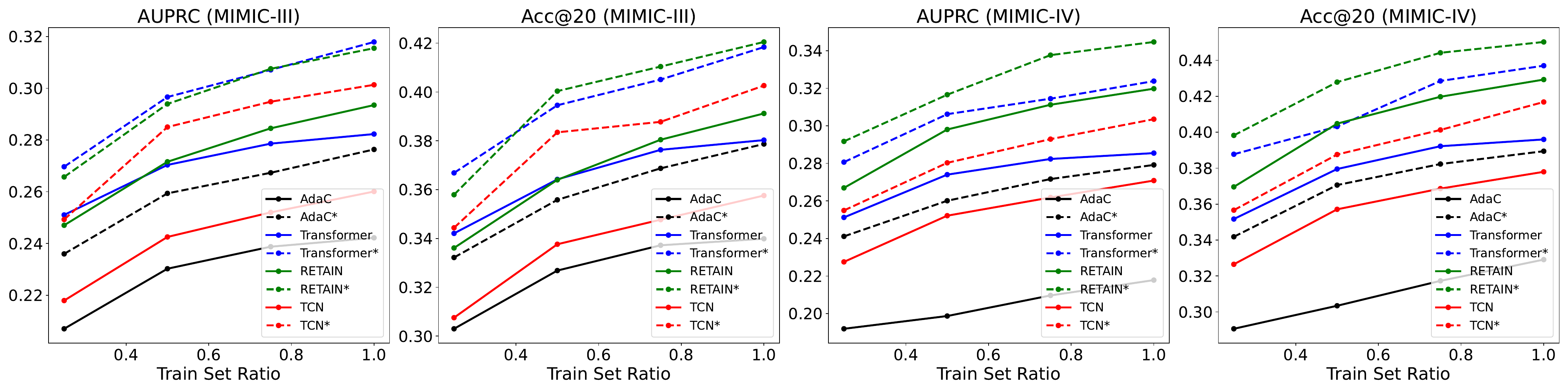}
\vspace{-0.2cm}
\caption{Performance evaluation across different training set sizes using the MIMIC-III and MIMIC-IV datasets. An asterisk (*) next to each encoder indicates the integration of \model{} to that model, e.g., Transformer* = Transformer + \model{}.}
\label{fig:enhancement_rare_code}
\end{center}
\vspace{-0.2cm}
\end{figure*}

As shown in Table \ref{tab:baseline_comparison_w_rare_codes}, we conduct an ablation study evaluating \model{} by removing key components:  
(1) \textbf{w/o leaf-HMP}: no horizontal message passing at the last level,  
(2) \textbf{w/o parent-HMP}: no horizontal message passing at parent levels,  
(3) \textbf{w/o HMP}: no horizontal message passing at any level,  
(4) \textbf{w/o HGIP}: no hierarchical graph information propagation in vertical message passing,  
(5) \textbf{w/o LLM}: random initialization for concept embeddings (removing LLM initialization).  
All ablated versions showed performance degradation. The drop in \textbf{w/o leaf-HMP} highlights the importance of ontology fusion at the last (leaf/child) level, while \textbf{w/o parent-HMP} exhibits an even greater decline, emphasizing the significance of ontology fusion at coarser-grained (parent) levels, which has been largely overlooked in prior studies. The largest drop occurs in \textbf{w/o HMP}, indicating the critical role of multi-ontology connections across all levels. The decline in \textbf{w/o HGIP} underscores the impact of vertical propagation through HGIP in structured hierarchical subgraphs. Finally, the performance drop in \textbf{w/o LLM} highlights the effectiveness of LLM-based embedding initialization, which injects clinical knowledge into the model, provides a strong training foundation, and guides it toward better performance.

To further assess the impact of ontology integration (fusion) on concept representation learning and overall performance, we train our model using different combinations of concept types (diagnosis (dx), drugs (rx), and procedures (px)), each time drop one concept type to compare the performance with when training on all.  Also, we repeat the same experiments with \model{} w/o HMP, an ablated version of our model which HMP is removed, meaning ontological multi-level integration is disabled. Based on Table \ref{tab:onto_colab}, we draw the following conclusions:  (1) Removing ontology integration (i.e., HMP) in any combination cases (comparing results on the left and right sides of the table) leads to a noticeable drop in performance. This highlights the importance of HMP in enabling multi-ontology integration and capturing both heterogeneous and homogeneous code interactions across all hierarchical levels. (2) Dropping any concept type (ontology) results in a performance decline when ontology integration is enabled (right-side of the table), with dx having the highest impact and px the least for diagnosis prediction. However, when ontology integration is removed (left-side of the table), this is not always the case. For example, training on (dx, rx) yields better results than (dx, rx, px) in the absence of ontology integration. This occurs because px information is significantly sparser than dx and rx, and when learned independently, it not only fails to contribute to diagnosis prediction but even degrades performance.  Without ontology integration, each ontology is forced to be learned independently, making it insufficient to extract useful information for diagnosis prediction. However, with HMP, px concepts are connected to dx and rx across all hierarchical levels, facilitating pattern extraction in concept representation learning.

\vspace{-0.4cm}
\subsection{RQ4: Prompt Analysis}
As shown in Table \ref{tab:llm_prompting_analysis}, we also experiment with different prompting strategies to determine the most effective information to include. The best results are achieved when the prompt includes the code type and name (e.g., ICD-9 Diagnosis 250.7), its corresponding descriptive concept (e.g., Diabetes with peripheral circulatory disorders), its parent codes and concepts, and the task name (e.g., diagnosis prediction). Notably, incorporating child information for parent codes slightly decreases performance, likely due to excessive prompt length, as each parent has many children. In contrast, adding parent information maintains a manageable prompt length since each code has only one parent per level. The second row of the table shows that adding irrelevant noise to the prompt degrades performance. In the last row of the table, we freeze the LLM embedding initialization and trained the model with fixed embeddings. The poor performance demonstrates that raw LLM embeddings without further refinement are ineffective. Lastly, we noticed that the LLM initialization technique accelerates training convergence, requiring about \( \sim 60 \) epochs compared to \( \sim 250 \) with random initialization.

\vspace{-2mm}
\subsection{RQ5: Data Insufficiency Analysis}

To assess the robustness of \model{} under data insufficiency, we vary the size of the training dataset to simulate limited data scenarios. As shown in Figure~\ref{fig:basline_trainset_size}, our model consistently outperforms its components in data-scarce settings, demonstrating its effectiveness. Furthermore, in the  Plug-in Enhancement setting, Figure~\ref{fig:enhancement_rare_code} shows that even with reduced training data, our model significantly enhances the performance of EHR models.

\subsection{RQ6: Case Study Analysis}
\label{app:case study}

\begin{figure}[t]
\begin{center}
\includegraphics[width=0.65\linewidth]{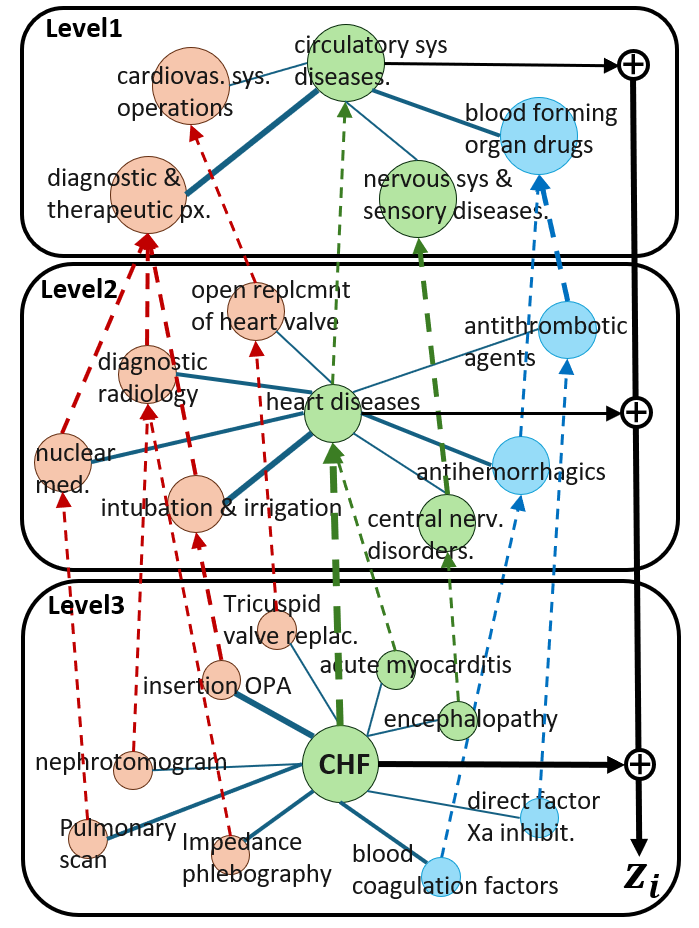}
\caption{Case study: Example of an extracted sub-KG for ICD-9 medical concept 428.0, representing ``Congestive Heart Failure (CHF)'', which demonstrates how \model{} learns representations through dual-axis message passing.} 
\label{fig:case_study}
\end{center}
\vspace{-5mm}
\end{figure}

Figure \ref{fig:case_study} illustrates how \model{} learns the representation of the ICD-9 code 428.0, which denotes ``Congestive heart failure, unspecified'' (CHF), through a dual-axis message passing paradigm for rich medical concept representations. Vertically, \model{} retrieves all ancestors of this code across levels, and horizontally, it gathers co-occurring codes (red for procedures, green for diagnosis, and blue for drugs) for each parent and the target code across all ontologies. The figure shows the extracted sub-KG for ICD-9 code 428.0 within the Meta-KG. Representation learning begins by first initializing each node embedding using LLM prompting and dense embedding retrieval. For example, for ICD-9 428.0 code, the prompt is << For the task of diagnosis prediction, provide a semantic representation for ICD-9 code 428.0 representing ``Congestive heart failure, unspecified". It falls under the broader ICD-9 categories of `420-429.99' (other forms of heart disease) and `390-459.99' (disease of the circulatory system)>>. \model{} then performs horizontal propagation, applying graph attention to aggregate information from neighboring nodes across all levels. Next, the HGIP propagates information upward, updating each parent node using its children's embeddings via the sequence of graph attention operators. Finally, the node embedding is refined through a convex combination of its own representation and those of its ancestors. 

Larger circles indicate higher-level codes, representing more general concepts compared to lower-level codes. Within each hierarchical level, solid blue lines connect the codes, representing co-occurrence-driven edges for horizontal message passing. Dashed lines illustrate parent-child relationships for bottom-up vertical message passing (HGIP), with color coding (red for procedures, green for diagnoses, and blue for drugs). Parent-child edges for top-down GRAM propagation are depicted using solid black lines. The weights for all graph edges during horizontal and vertical propagation are learned through attention techniques, with edge thickness in the figure indicating the relative attention assigned to each edge.

\section{Related Work}
\label{related_work}

\textbf{EHR Predictive Models.}  The widespread adoption of EHRs has enabled the development of machine learning models for healthcare prediction, starting with sequential models \citep{choi2016doctor, ashfaq2019readmission}, followed by attention-based models \citep{choi2016retain, hu2025recurrent}, transformer-based approaches \citep{li2020behrt, choi2020learning, nayebi2023contrastive}, and recently, graph neural networks \citep{su2020gate, lu2021self, xu2022counterfactual, yang2023molerec, poulain2024graph}.

\noindent\textbf{Ontology-Driven Concept Representation.}
These works aim to enhance medical concept representation learning by augmenting structured EHR data (sequence of discrete medical codes) with hierarchical medical ontology, without using any other data modalities (e.g., unstructured clinical text, numerical measurements). We compare our method with these baselines in our experiments. For instance, GRAM \citep{choi2017gram} leverages the ontology hierarchy to represent a medical concept as a convex combination of itself and its ancestors. Building on GRAM,
MMORE \citep{song2019medical} enhances GRAM by enabling multiple representations for each parent concept, addressing discrepancies between EHR data and medical ontologies. KAME \citep{ma2018kame} further proposes incorporating ontology knowledge throughout the entire prediction process on top of code representation learning.  While GRAM-based methods improve performance, they limit expressiveness by treating ancestors as unordered. HAP \citep{zhang2020hierarchical} addresses this by propagating attention hierarchically, enabling top-down and bottom-up information sharing across the ontology. 
Further, G-BERT \citep{shang2019pre} combines GNNs on ontology and BERT for medical code representation. Later, ADORE \citep{cheong2023adaptive} utilizes the relational ontology SNOMED to integrate multi-source medical codes, which was ignored in previous studies. However, it suffers from loss of information in code conversion from ICD (hierarchical) to SNOMED (relational), especially for medication codes. Also, KAMPNet \citep{an2023kampnet} employs graph contrastive learning for effective EHR representation, but first fails to fully account for the hierarchical order of node calculation, and second, it only captures the interaction of multi-source codes (diagnosis, medication) in the last level of ontologies.

\noindent\textbf{Multi-Source-Augmented EHR Representation} \citep{lu2021collaborative, lu2024clinicalrag, safaeipour2024semantic, jiang2024reasoning}.  GCL \citep{lu2021collaborative} is a collaborative graph learning model that jointly learns patient and disease representations, incorporating unstructured text with attention regulation. RAM-EHR \citep{xu2024ram} improves EHR predictions by retrieving external textual medical knowledge from multiple online sources, augmenting the local model co-trained with consistency regularization.  EMERGE \citep{10.1145/3627673.3679582}, a RAG framework, enhances multimodal health representations by encoding clinical notes with an LLM and time-series data with a GRU. KARE \citep{jiang2024reasoning} integrates structured EHR codes and unstructured biomedical text via multi-source knowledge graphs and community-level retrieval. It enhances clinical prediction by augmenting patient context and fine-tuning LLMs with reasoning chains. ClinicalRAG \citep{lu2024clinicalrag} uses a multi-agent RAG pipeline to integrate structured (e.g., knowledge graphs) and unstructured (e.g., online text) medical knowledge into LLMs, improving its diagnostic accuracy. In experiments, we do not compare with such models that incorporate external descriptive texts, as they operate in a multimodal setting and focus on general patient representation. In contrast, our model targets concept-level representation using only structured code.


\vspace{-0.3cm}
\section{Conclusion}

We introduced LINKO, a framework for LLM-augmented multi-ontology integration via dual-axis knowledge propagation, designed to enhance medical concept representation in EHR models. \model{} extracts cross-ontology relationships through message passing in two dimensions: vertical and horizontal, and initializes concept embeddings with graph-retrieval-augmented LLM prompting. By enhancing medical concept representations, \model{} achieves superior performance on downstream tasks compared to other medical concept encoder baselines. Additionally, we showcase the robustness of \model{} in data-limited scenarios and validate its plug-in compatibility to enhance existing healthcare models.

\section{GenAI Usage Disclosure}
We used generative AI tools (e.g., ChatGPT) solely for the purpose of grammatical and stylistic editing of the manuscript. No content generation, data analysis, or experimental design was performed using these tools.

\bibliographystyle{ACM-Reference-Format}
\balance
\bibliography{sample-base}

\end{document}